\documentclass[11pt]{article}
\usepackage{acl2016}
\usepackage{times}
\usepackage{url}
\usepackage{latexsym}
\usepackage{epsfig}
\usepackage{graphicx}
\graphicspath{ {figures/} }
\usepackage{array}
\usepackage{amsmath}
\usepackage{amssymb}
\usepackage{algpseudocode}
\usepackage{algorithm}
\usepackage{algorithmicx}
\usepackage{caption,booktabs}
\usepackage{bbm}
\usepackage{amsmath}
\usepackage[table]{xcolor}
\usepackage{float}
\usepackage{color}

\newcommand{\argmax}{\operatornamewithlimits{argmax}}
\aclfinalcopy % Uncomment this line for the final submission
\title{Multimodal Pivots for Image Caption Translation}

\author{Julian Hitschler \and Shigehiko Schamoni \\
  Computational Linguistics \\
  Heidelberg University \\
  69120 Heidelberg, Germany \\
  {\tt \small \{hitschler,schamoni\}@cl.uni-heidelberg.de} \\
  \And
  Stefan Riezler \\
  Computational Linguistics \& IWR \\
  Heidelberg University \\
  69120 Heidelberg, Germany \\
 {\tt \small riezler@cl.uni-heidelberg.de} }
      
\date{}

\begin{document}

\maketitle

\begin{abstract}
  We present an approach to improve statistical machine translation of image descriptions by multimodal pivots defined in visual space. The key idea is to perform image retrieval over a database of images that are captioned in the target language, and use the captions of the most similar images for crosslingual reranking of translation outputs. Our approach does not depend on the availability of large amounts of in-domain parallel data, but only relies on available large datasets of monolingually captioned images, and on state-of-the-art convolutional neural networks to compute image similarities. Our experimental evaluation shows improvements of 1 BLEU point over strong baselines.
\end{abstract}

\section{Introduction}
Multimodal data consisting of images and natural language descriptions (henceforth called \emph{captions}) are an abundant source of information that has led to a recent surge in research integrating language and vision. Recently, the aspect of multilinguality has been added to multimodal language processing in a shared task at the WMT16 conference.\footnote{\url{http://www.statmt.org/wmt16/multimodal-task.html}} There is clearly also a practical demand for multilingual image captions, e.g., automatic translation of descriptions of art works would allow access to digitized art catalogues across language barriers and is thus of social and cultural interest; multilingual product descriptions are of high commercial interest since they would allow to widen e-commerce transactions automatically to international markets.
However, while datasets of images and monolingual captions already include millions of tuples \cite{FerraroETAL:15}, the largest multilingual datasets of images and captions known to the authors contain 20,000 \cite{GrubingerETAL:06} or 30,000\footnote{The dataset used at the WMT16 shared task is based on translations of Flickr30K captions \cite{Flickr30k}.} triples of images with German and English descriptions. 

In this paper, we want to address the problem of multilingual captioning from the perspective of statistical machine translation (SMT). In contrast to prior work on generating captions directly from images (\newcite{KulkarniETAL:11}, \newcite{KarpathyFeiFei:15}, \newcite{VinyalsETAL:15}, \emph{inter alia}), our goal is to integrate visual information into an SMT pipeline. Visual context provides orthogonal information that is free of the ambiguities of natural language, therefore it serves to disambiguate and to guide the translation process by grounding the translation of a source caption in the accompanying image. Since datasets consisting of source language captions, images, and target language captions are not available in large quantities, we would instead like to utilize large datasets of images and target-side monolingual captions to improve SMT models trained on modest amounts of parallel captions.

Let the task of \emph{caption translation} be defined as follows: For production of a target caption $e_i$ of an image $i$, a system may use as input an image caption for image $i$ in the source language $f_i$, as well as the image $i$ itself. The system may safely assume that $f_i$ is relevant to $i$, i.e., the identification of relevant captions for $i$ \cite{Hodosh2013}
is not itself part of the task of caption translation. In contrast to the inference problem of finding $\hat{e} = \argmax_e p(e|f)$ in text-based SMT, multimodal caption translation allows to take into consideration $i$ as well as $f_i$ in finding $\hat{e}_i$: 
$$\hat{e}_i = \argmax_{e_i} p(e_i|f_i,i)$$

In this paper, we approach caption translation by a general crosslingual reranking framework where for a given pair of source caption and image, monolingual captions in the target language are used to rerank the output of the SMT system. We present two approaches to retrieve target language captions for reranking by pivoting on images that are similar to the input image. One approach calculates image similarity based deep convolutional neural network (CNN) representations. Another approach calculates similarity in visual space by comparing manually annotated object categories. We compare the multimodal pivot approaches to reranking approaches that are based on text only, and to standard SMT baselines trained on parallel data. Compared to a strong baseline trained on 29,000 parallel caption data, we find improvements of over 1 BLEU point for reranking based on visual pivots. Notably, our reranking approach does not rely on large amounts of in-domain parallel data which are not available in practical scenarios such as e-commerce localization. However, in such scenarios, monolingual product descriptions are naturally given in large amounts, thus our work is a promising pilot study towards real-world caption translation.

\section{Related Work}
Caption generation from images alone has only recently come into the scope of realistically solvable problems in image processing (\newcite{KulkarniETAL:11}, \newcite{KarpathyFeiFei:15}, \newcite{VinyalsETAL:15}, \emph{inter alia}). Recent approaches also employ reranking of image captions by measuring similarity between image and text using deep representations \cite{FangETAL:15}. The tool of choice in these works are neural networks whose deep representations have greatly increased the quality of feature representations of images, enabling robust and semantically salient analysis of image content. We rely on the CNN framework \cite{Socher2014,SimonyanZisserman:15} to solve semantic classification and disambiguation tasks in NLP with the help of supervision signals from visual feedback. However, we consider image captioning as a different task than caption translation since it is not given the information of the source language string. Therefore we do not compare our work to caption generation models.

In the area of SMT, \newcite{Waeschle2015} presented a framework for integrating a large, in-domain, target-side monolingual corpus into machine translation by making use of techniques from crosslingual information retrieval. The intuition behind their approach is to generate one or several translation hypotheses using an SMT system, which act as queries to find matching, semantically similar sentences in the target side corpus. These can in turn be used as templates for refinement of the translation hypotheses, with the overall effect of improving translation quality. Our work can be seen as an extension of this method, with visual similarity feedback as additional constraint on the crosslingual retrieval model.
\newcite{Calixto2012} suggest using images as supplementary context information for statistical machine translation. They cite examples from the news domain where visual context could potentially be helpful in the disambiguation aspect of SMT and discuss possible features and distance metrics for context images, but do not report experiments involving a full SMT pipeline using visual context.
In parallel to our work, \newcite{DBLP:journals/corr/ElliottFH15} addressed the problem of caption translation from the perspective of neural machine translation.\footnote{We replicated the results of \newcite{DBLP:journals/corr/ElliottFH15} on the IAPR TC-12 data. However, we decided to not include their model as baseline in this paper since we found our hierarchical phrase-based baselines to yield considerably better results on IAPR TC-12 as well as on MS COCO.}
Their approach uses a model which is considerably more involved than ours and relies exclusively on the availability of parallel captions as training data. Both approaches crucially rely on neural networks, where they use a visually enriched neural encoder-decoder SMT approach, while we follow a retrieval paradigm for caption translation, using CNNs to compute similarity in visual space.

Integration of multimodal information into NLP problems has been another active area of recent research. For example, \newcite{silberer-lapata:2014:P14-1} show that distributional word embeddings grounded in visual representations outperform competitive baselines on term similarity scoring and word categorization tasks. The orthogonality of visual feedback has previously been exploited in a multilingual setting by \newcite{kiela-vulic-clark:2015:EMNLP} (relying on previous work by \newcite{Bergsma:11}), who induce a bilingual lexicon using term-specific multimodal representations obtained by querying the Google image search engine.\footnote{\url{https://images.google.com/}} \newcite{funaki-nakayama:2015:EMNLP} use visual similarity for crosslingual document retrieval in a multimodal and bilingual vector space obtained by generalized canonical correlation analysis, greatly reducing the need for parallel training data.  The common element is that CNN-based visual similarity information is used as  a ``hub'' \cite{funaki-nakayama:2015:EMNLP} or pivot connecting corpora in two natural languages which lack direct parallelism, a strategy which we apply to the problem of caption translation.

\section{Models}
\subsection{Overview}

\begin{figure}[t]
    \centering 
    \includegraphics[scale=0.23]{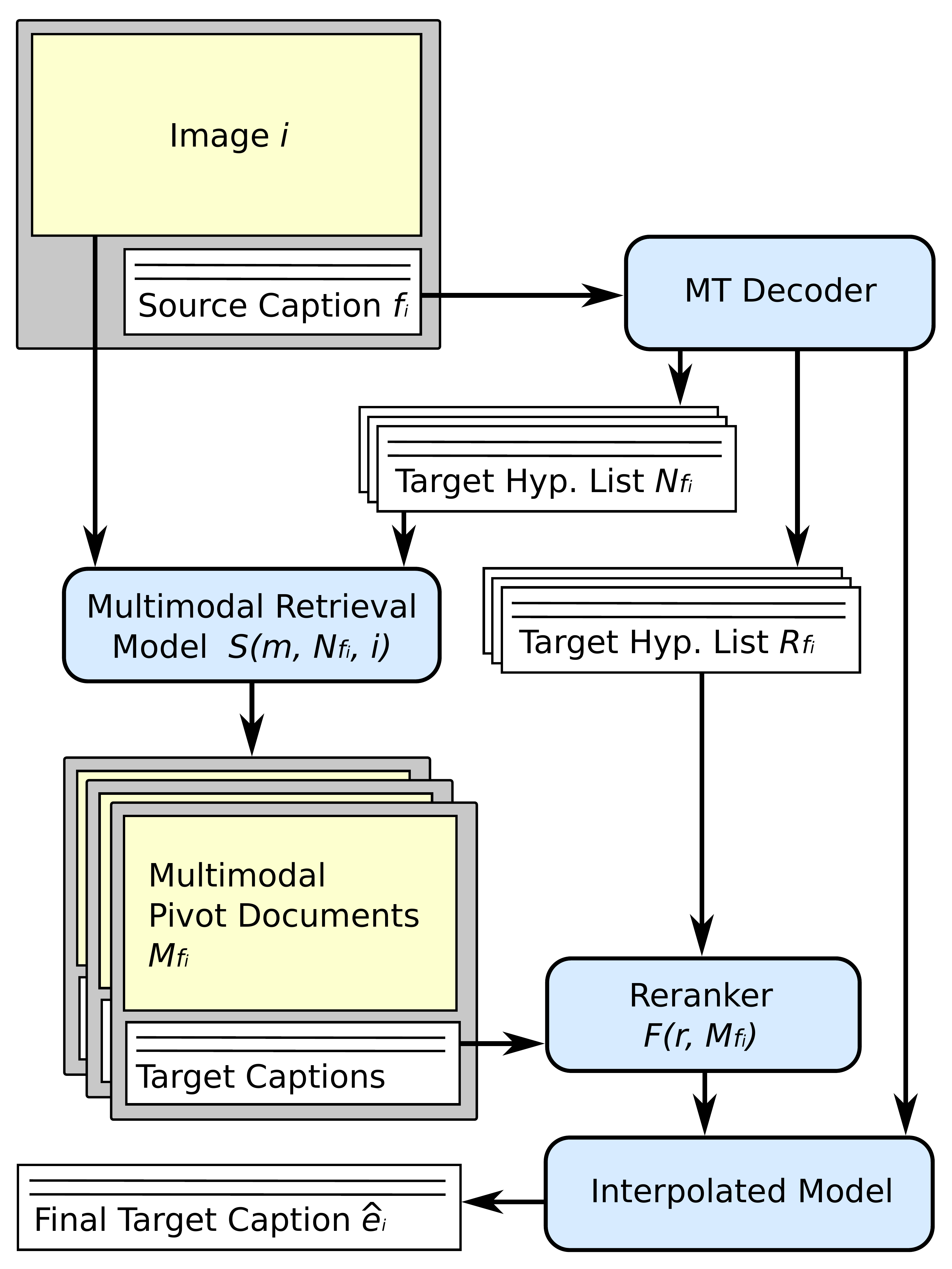}
    \caption{\label{figure:pipeline} Overview of model architecture.}
\end{figure}

Following the basic approach set out by \newcite{Waeschle2015}, we use a crosslingual retrieval model to find sentences in a target language document collection $C$, and use these to rerank target language translations $e$ of a source caption $f$.

The systems described in our work differ from that of \newcite{Waeschle2015} in a number of aspects. Instead of a two-step architecture of coarse-grained and fine-grained retrieval, our system uses relevance scoring functions for retrieval of matches in the document collection $C$, and for reranking of translation candidates that are based on inverse document frequency of terms \cite{SpaerckJones1972} and represent variants of the popular TF-IDF relevance measure.

A schematic overview of our approach is given in Figure \ref{figure:pipeline}. It consists of the following components:

\begin{description}
 \item[Input:] Source caption $f_i$, image $i$, target-side collection $C$ of image-captions pairs
 \item[Translation:] Generate unique list $N_{f_i}$ of $k_n$-best translations, generate unique list $R_{f_i}$ of $k_r$-best list of translations\footnote{In practice, the first hypothesis list may be reused. We distinguish between the two hypothesis lists $N_{f_i}$ and $R_{f_i}$ for notational clarity since in general, the two hypothesis lists need not be of equal length.} using MT decoder
   \item[Multimodal retrieval:] For list of translations $N_{f_i}$, find set $M_{f_i}$ of $k_m$-most relevant pairs of images and captions in a target-side collection $C$, using a heuristic relevance scoring function $S(m, N_{f_i},i), m \in C$
   \item[Crosslingual reranking:] Use list $M_{f_i}$ of image-caption pairs to rerank list of translations $R_{f_i}$, applying relevance scoring function $F(r, M_{f_i})$ to all $r \in R_{f_i}$
     \item[Output:] Determine best translation hypothesis $\hat{e}_i$  by interpolating decoder score $d_r$ for a hypothesis $r \in R_{f_i}$ with its relevance score $F(r, M_{f_i})$ with weight $\lambda$ s.t. $$ \hat{e}_i = \argmax_{r \in R_{f_i}} d_r + \lambda \cdot F(r, M_{f_i})$$
  \end{description}

The central concept is the scoring function $S(m, N_{f_i},i)$ which defines three variants of target-side retrieval (TSR), all of which make use of the procedure outlined above. 
In the baseline text-based reranking model (TSR-TXT), we use relevance scoring function $S_{TXT}$. This function is purely text-based and does not make use of multimodal context information (as such, it comes closest to the models used for target-side retrieval in \newcite{Waeschle2015}). In the retrieval model enhanced by visual information from a deep convolutional neural network (TSR-CNN), the scoring function $S_{CNN}$ incorporates a textual relevance score with visual similarity information extracted from the neural network. Finally, we evaluate these models against a relevance score based on human object-category annotations (TSR-HCA), using the scoring function $S_{HCA}$. This function makes use of the object annotations available for the MS COCO corpus \cite{mscoco} to give an indication of the effectiveness of our automatically extracted visual similarity metric. The three models are discussed in detail below.

\subsection{Target Side Retrieval Models}

\paragraph*{Text-Based Target Side Retrieval.}
\label{section:text_retrieval}

In the TSR-TXT retrieval scenario, a match candidate $m \in C$ is scored in the following way:

\begin{align*}
&S_{TXT}(m, N_{f_i}) = \\
& Z_m\sum_{n \in N_{f_i}}\sum_{w_n \in tok(n)}\sum_{w_m \in typ(m)} \delta(w_m,w_n)  idf(w_m), \\
\end{align*}
where $\delta$ is the Kronecker $\delta$-function, $N_{f_i}$ is the set of the $k_n$-best translation hypotheses for a source caption $f_i$ of image $i$ by decoder score, $typ(a)$ is a function yielding the set of types (unique tokens) contained in a caption $a$,\footnote{The choice for per-type scoring of reference captions was primarily driven by performance considerations. Since captions rarely contain repetitions of low-frequency terms, this has very little effect in practice, other than to mitigate the influence of stopwords.} $tok(a)$ is a function yielding the tokens of caption $a$, $idf(w)$ is the inverse document frequency \cite{SpaerckJones1972} of term $w$, and $Z_m = \frac{1}{|typ(m)|}$ is a normalization term introduced in order to avoid biasing the system towards long match candidates containing many low-frequency terms. Term frequencies were computed on monolingual data from Europarl \cite{Koehn2005} and the News Commentary and News Discussions English datasets provided for the WMT15 workshop.\footnote{\url{http://www.statmt.org/wmt15/translation-task.html}} Note that in this model, information from the image $i$ is not used. 

\paragraph*{Multimodal Target Side Retrieval using CNNs.}

In the TSR-CNN scenario, we supplement the textual target-side TSR model with visual similarity information from a deep convolutional neural network. We formalize this by introduction of the positive-semidefinite distance function $v(i_x, i_y) \rightarrow [0,\infty)$ for images $i_x$, $i_y$ (smaller values indicating more similar images). The relevance scoring function $S_{CNN}$ used in this model takes the following form:
\begin{align*}
  &S_{CNN}(m, N_{f_i}, i) \\
  &=
  \begin{cases}
    S_{TXT}(m, N_{f_i}) e^{ - bv(i_m,i)},  & v(i_m,i) < d \\
    0 & otherwise,\\
  \end{cases}
\end{align*}
where $i_m$ is the image to which the caption $m$ refers and $d$ is a cutoff maximum distance, above which match candidates are considered irrelevant, and $b$ is a weight term which controls the impact of the visual distance score $v(i_m, i)$ on the overall score.\footnote{The value of $b =0.01$ was found on development data and kept constant throughout the experiments.}

Our visual distance measure $v$ was computed using the VGG16 deep convolutional model of \newcite{SimonyanZisserman:15}, which was pre-trained on ImageNet \cite{DBLP:journals/corr/RussakovskyDSKSMHKKBBF14}. We extracted feature values for all input and reference images from the penultimate fully-connected layer ({\tt fc7}) of the model and computed the Euclidean distance between feature vectors of images. If no neighboring images fell within distance $d$, the text-based retrieval procedure $S_{TXT}$ was used as a fallback strategy, which occurred 47 out of 500 times on our test data.

\paragraph*{Target Side Retrieval by Human Category Annotations.}
\label{section:image_retrieval_human}

For contrastive purposes, we evaluated a TSR-HCA retrieval model which makes use of the human object category annotations for MS COCO. Each image in the MS COCO corpus is annotated with object polygons classified into 91 categories of common objects. In this scenario, a match candidate $m$ is scored in the following way:
\begin{align*}
  &S_{HCA}(m, N_{f_i}, i) \\
  &= \delta(cat(i_m),cat(i)) S_{TXT}(m, N_{f_i}),
  \end{align*}
 where $cat(i)$ returns the set of object categories with which image $i$ is annotated. The amounts to enforcing a strict match between the category annotations of $i$ and the reference image $i_m$, thus pre-filtering the $S_{TXT}$ scoring to captions for images with strict category match.\footnote{Attempts to relax this strict matching criterion led to strong performance degradation on the development test set.} In cases where $i$ was annotated with a unique set of object categories and thus no match candidates with nonzero scores were returned by $S_{HCA}$, $S_{TXT}$ was used as a fallback strategy, which occurred 77 out of 500 times on our test data.

\subsection{Translation Candidate Re-scoring}
\label{section:rescoring}

The relevance score $F(r, M_{f_i})$ used in the reranking model was computed in the following way for all three models:

\begin{align*}
&F(r, M_{f_i}) =\\
& Z_{M_{f_i}} \sum_{m \in M_{f_i}}\sum_{w_m \in typ(m)}\sum_{w_r \in tok(r)}\delta(w_m, w_r)idf(w_m)
\end{align*}
with normalization term
\[
Z_{M_{f_i}} = (\sum_{m \in M_{f_i}}|tok(m)|)^{-1},
\]
where $r$ is a translation candidate and $M_{f_i}$ is a list of $k_m$-top target side retrieval matches. Because the model should return a score that is reflective of the relevance of $r$ with respect to $M_{f_i}$, irrespective of the length of $M_{f_i}$, normalization with respect to the token count of $M_{f_i}$ is necessary. The term $Z_{M_{f_i}}$ serves this purpose.

\section{Experiments}
\label{chapter:experiments}

\subsection{Bilingual Image-Caption Data}

We constructed a German-English parallel dataset based on the MS COCO image corpus \cite{mscoco}. 1,000 images were selected at random from the 2014 training section\footnote{We constructed our parallel dataset using only the training rather than the validation section of MS COCO so as to keep the latter pristine for future work based on this research.} and, in a second step, one of their five English captions was chosen randomly. This caption was then translated into German by a native German speaker. Note that our experiments were performed with German as the source and English as the target language, therefore, our reference data was not produced by a single speaker but reflects the heterogeneity of the MS COCO dataset at large. The data was split into a development set of 250 captions, a development test set of 250 captions for testing work in progress, and a test set of 500 captions. For our retrieval experiments, we used only the images and captions that were not included in the development, development test or test data, a total of 81,822 images with 5 English captions per image. All data was tokenized and converted to lower case using the {\tt cdec}\footnote{\url{https://github.com/redpony/cdec}} utilities {\tt tokenized-anything.pl} and {\tt lowercase.pl}. For the German data, we performed compound-splitting using the method described by \newcite{Dyer2009}, as implemented by the {\tt cdec} utility {\tt compound-split.pl}. Table \ref{table:img_data} gives an overview of the dataset. Our parallel development, development test and test data is publicly available.\footnote{\url{www.cl.uni-heidelberg.de/decoco/}}

\begin{table}[t]
\begin{center}
\resizebox{\columnwidth}{!}{
\begin{tabular}{lccc}

\toprule
  Section & Images  & Captions &   Languages \\
\midrule
  DEV & 250 & 250 &   DE-EN \\
  DEVTEST & 250 & 250 &   DE-EN \\
  TEST & 500 & 500 &   DE-EN \\
  RETRIEVAL ($C$) & 81,822 & 409,110 &  EN \\
\bottomrule
\end{tabular}
}
\end{center}
\caption{\label{table:img_data} Number of images and sentences in MS COCO image and caption data used in experiments.}
\end{table}

\subsection{Translation Baselines}
\label{section:translation_system}
We compare our approach to two baseline machine translation systems, one trained on out-of-domain data exclusively and one domain-adapted system. Table \ref{table:mt_data} gives an overview of the training data for the machine translation systems. 

\paragraph*{Out-of-Domain Baseline.}

Our baseline SMT framework is hierarchical phrase-based translation using synchronous context free grammars \cite{Chiang2007}, as implemented by the {\tt cdec} decoder \cite{Dyer2010}. Data from the Europarl \cite{Koehn2005}, News Commentary and Common Crawl corpora \cite{Smith2013} as provided for the WMT15 workshop was used to train the translation model, with German as source and English as target language.

Like the retrieval dataset, training, development and test data was tokenized and converted to lower case, using the same {\tt cdec} tools. Sentences with lengths over 80 words in either the source or the target language were discarded before training. Source text compound splitting was performed using {\tt compound-split.pl}. Alignments were extracted bidirectionally using the {\tt fast-align} utility of {\tt cdec} and symmetrized with the {\tt atools} utility (also part of {\tt cdec}) using the {\tt grow-diag-final-and} symmetrization heuristic. The alignments were then used by the {\tt cdec} grammar extractor to extract a synchronous context free grammar from the parallel data. 

The target language model was trained on monolingual data from Europarl, as well as the News Crawl and News Discussions English datasets provided for the WMT15 workshop (the same data as was used for estimating term frequencies for the retrieval models) with the {\tt KenLM} toolkit \cite{Heafield-estimate,Heafield-kenlm}.\footnote{\url{https://kheafield.com/code/kenlm/}}

We optimized the parameters of the translation system for translation quality as measured by IBM BLEU \cite{Papineni2002} using the Margin Infused Relaxed Algorithm (MIRA) \cite{Crammer2003}. For tuning the translation models used for extraction of the hypothesis lists for final evaluation, MIRA was run for 20 iterations on the development set, and the best run was chosen for final testing.

\paragraph*{In-Domain Baseline.}

We also compared our models to a domain-adapted machine translation system. The domain-adapted system was identical to the out-of-domain system, except that it was supplied with additional parallel training data from the image caption domain. For this purpose, we used 29,000 parallel German-English image captions as provided for the WMT16 shared task on multimodal machine translation. The English captions in this dataset belong to the Flickr30k corpus \cite{Flickr30k} and are very similar to those of the MS COCO corpus. The German captions are expert translations. The English captions were also used as additional training data for the target-side language model. We generated $k_n$- and $k_r$-best lists of translation candidates using this in-domain baseline system. 

\begin{table}[t]
\begin{center}
\resizebox{\columnwidth}{!}{
\begin{tabular}{lccc}

\toprule
  Corpus & Sentences &  Languages & System\\
\midrule
 Europarl & 1,920,209 & DE-EN & O/I \\
 News Commentary & 216,190 & DE-EN & O/I \\
 Common Crawl & 2,399,123 & DE-EN & O/I \\
 Flickr30k WMT16 & 29,000 & DE-EN & I \\
\midrule
 Europarl & 2,218,201 & EN & O/I \\
 News Crawl & 28,127,448 & EN & O/I\\
 News Discussions & 57,803,684 & EN & O/I\\
 Flickr30k WMT16 & 29,000 & EN & I \\

\bottomrule
\end{tabular}
}
\end{center}
\caption{\label{table:mt_data} Parallel and monolingual data used for training machine translation systems. Sentence counts are given for raw data without pre-processing. O/I: both out-of-domain and in-domain system, I: in-domain system only.}
\end{table}

\subsection{Optimization of TSR Hyperparameters}
\label{section:retrieval_tuning}

For each of our retrieval models, we performed a step-wise exhaustive search of the hyperparameter space over the four system hyperparameters for IBM BLEU on the development set: The length of the $k_n$-best list the entries of which are used as queries for retrieval; the number of $k_m$-best-matching captions retrieved; the length of the final $k_r$-best list used in reranking; the interpolation weight $\lambda$ of the relevance score $F$ relative to the translation hypothesis log probability returned by the decoder. The parameter ranges to be explored were determined manually, by examining system output for prototypical examples. Table \ref{table:tuning} gives an overview over the hyperparameter values obtained. 

For TSR-CNN, we initially set the cutoff distance $d$ to 90.0, after manually inspecting sets of nearest neighbors returned for various maximum distance values. After optimization of retrieval parameters, we performed an exhaustive search from $d=80.0$ to $d=100.0$, with step size 1.0 on the development set, while keeping all other hyperparameters fixed, which confirmed out initial choice of $d=90.0$ as the optimal value. 

Explored parameter spaces were identical for all models and each model was evaluated on the test set using its own optimal configuration of hyperparameters. 

\begin{table}[t]
\begin{center}
%\resizebox{\columnwidth}{!}{
\begin{tabular}{lcccc}
\toprule
  Model & $k_n$ & $k_m$ & $k_r$ &$\lambda$ \\
\midrule
  TSR-TXT & $300$ & $500$ & $5$ & $5 \cdot 10^{4}$ \\
  TSR-CNN & $300$ & $300$ & $5$ & $70 \cdot 10^{4}$ \\
  TSR-HCA & $300$ & $500$ & $5$ & $10 \cdot 10^{4}$ \\
\bottomrule
\end{tabular}
%}
\end{center}
\caption{\label{table:tuning} Optimized hyperparameter values used in final evaluation.}
\end{table}

\subsection{Significance Testing}
\label{section:sigtesting}

Significance tests on the differences in translation quality were performed using the approximate randomization technique for measuring performance differences of machine translation systems described in \newcite{Riezler2005} and implemented by \newcite{Clark2011} as part of the {\tt Multeval} toolkit.\footnote{\url{https://github.com/jhclark/multeval}}

\subsection{Experimental Results}

Table \ref{tab:scores} summarizes the results for all models on an unseen test set of 500 captions. Domain adaptation led to a considerable improvement of +4.1 BLEU and large improvements in terms of METEOR and Translation Edit Rate (TER). We found that the target-side retrieval model enhanced with multimodal pivots from a deep convolutional neural network, TSR-CNN and TSR-HCA, consistently outperformed both the domain-adapted {\tt cdec} baseline, as well as the text-based target side retrieval model TSR-TXT. These models therefore achieve a performance gain which goes beyond the effect of generic domain-adaptation. The gain in performance for TSR-CNN and TSR-HCA was significant at $p<0.05$ for BLEU, METEOR, and TER. For all evaluation metrics, the difference between TSR-CNN and TSR-HCA was not significant, demonstrating that retrieval using our CNN-derived distance metric could match retrieval based the human object category annotations. 

\begin{table}[t]
\begin{center}
\resizebox{\columnwidth}{!}{
\begin{tabular}{llllll}
\toprule
 \bf System & BLEU $\uparrow$ & \bf $p_{c}$ & \bf $p_t$ & \bf $p_d$ & \bf $p_o$ \\
\midrule
 {\tt cdec} out-dom. & 25.5 & & & &  \\ 
 {\tt cdec} in-dom. & 29.6 & & & & 0.00 \\ 
  TSR-TXT & 29.7 & & & 0.45 & 0.00 \\
  TSR-CNN & \textbf{30.6} & & 0.04 & 0.02 & 0.00 \\
  TSR-HCA & \textbf{30.3} & 0.42 & 0.01 & 0.00 & 0.00 \\
\midrule
 \bf System & METEOR $\uparrow$ & \bf $p_{c}$ & \bf $p_t$ & \bf $p_d$ & \bf $p_o$ \\
\midrule
 {\tt cdec} out-dom. & 31.7 & & & & \\ 
 {\tt cdec} in-dom. & 34.0 & & & & 0.00 \\ 
  TSR-TXT & 34.1 & & & 0.41 & 0.00 \\
  TSR-CNN & \textbf{34.7} & & 0.00 & 0.00 & 0.00 \\
  TSR-HCA & \textbf{34.4} & 0.09 & 0.00 & 0.00 & 0.00 \\
\midrule
 \bf System & TER $\downarrow$ & \bf $p_{c}$ & \bf $p_t$ & \bf $p_d$ & \bf $p_o$ \\
\midrule
 {\tt cdec} out-dom. & 49.3 & & & & \\ 
 {\tt cdec} in-dom. & 46.1 & & & & 0.00 \\ 
  TSR-TXT & 45.8 & & & 0.12 & 0.00 \\
  TSR-CNN & \textbf{45.1} & & 0.03 & 0.00 & 0.00 \\
  TSR-HCA & \textbf{45.3} & 0.34 & 0.02 & 0.00 & 0.00 \\
\bottomrule
\end{tabular}
}
\end{center}
\caption{\label{tab:scores} Metric scores for all systems and their significance levels as reported by {\tt Multeval}. $p_o$-values are relative to the {\tt cdec} out-of-domain baseline, $p_d$-values are relative to the {\tt cdec} in-domain baseline, $p_t$-values are relative to TSR-TXT and $p_c$-values are relative to TSR-CNN. %Note that because of differing sample variances, higher differences in an evaluation score do not imply higher confidence levels, nor vice versa.
Best results are reported in \textbf{bold} face.\footnotemark } 
\end{table}
\footnotetext{A baseline for which a random hypothesis was chosen from the top-5 candidates of the in-domain system lies between the other two baseline systems: 27.5\,/\,33.3\,/\,47.7 (BLEU\,/\,METEOR\,/\,TER). } 

The text-based retrieval baseline TSR-TXT never significantly outperformed the in-domain {\tt cdec} baseline, but there were slight nominal improvements in terms of BLEU, METEOR and TER. This finding is actually consistent with \newcite{Waeschle2015} who report performance gains for text-based, target side retrieval models only on highly technical, narrow-domain corpora and even report performance degradation on medium-diversity corpora such as Europarl. Our experiments show that it is the addition of visual similarity information by incorporation of multimodal pivots into the image-enhanced models TSR-CNN and TSR-HCA which makes such techniques effective on MS COCO, thus upholding our hypothesis that visual information can be exploited for improvement of caption translation.

\begin{figure}[t]
    \centering 
    \includegraphics[scale=.6]{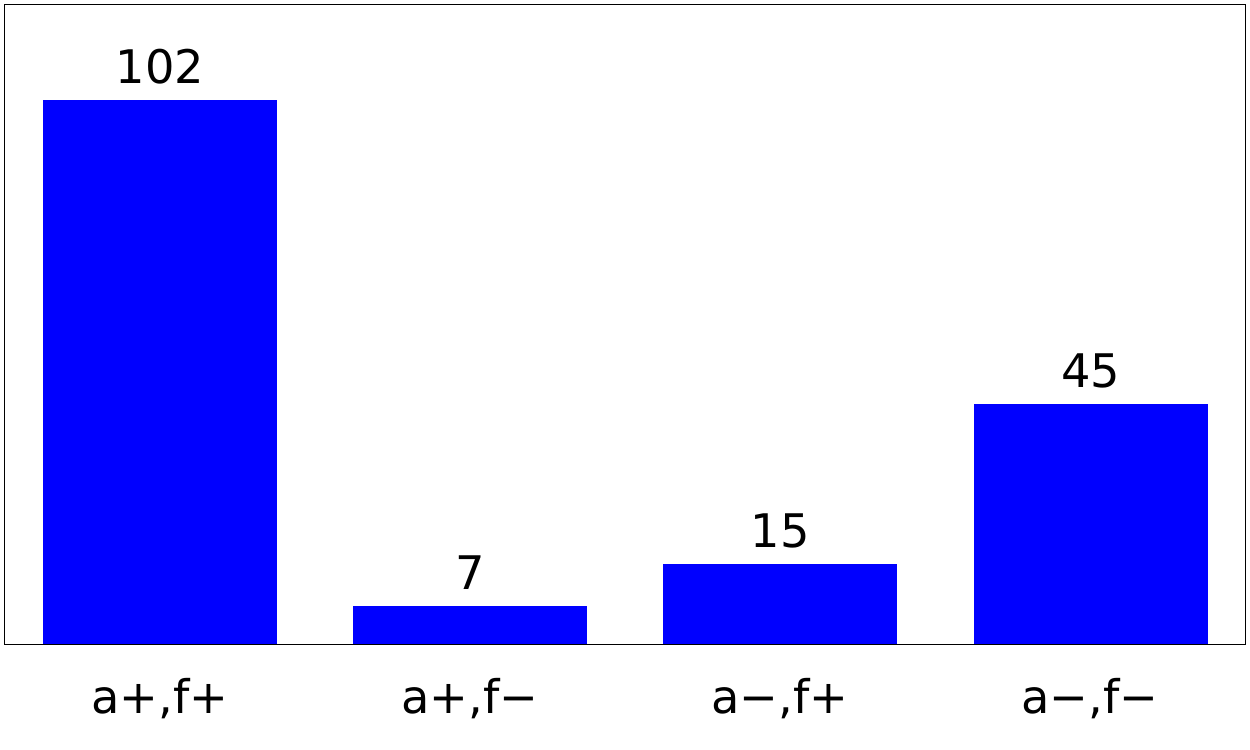}
    \caption{\label{figure:blindtest} Results of the human pairwise preference ranking experiment, given as the joint distribution of both rankings:
    $a+$ denotes preference for TSR-CNN in terms of accuracy, $f+$ in terms of fluency;
    $a-$ denotes preference for the in-domain baseline in terms of accuracy, $f-$ in terms of fluency. }
\end{figure}

\subsection{Human Evaluation}

The in-domain baseline and TSR-CNN differed in their output in 169 out of 500 cases on the test set. These 169 cases were presented to a human judge alongside the German source captions in a double-blinded pairwise preference ranking experiment. The order of presentation was randomized for the two systems. The judge was asked to rank fluency and accuracy of the translations independently. The results are given in Figure \ref{figure:blindtest}. Overall, there was a clear preference for the output of TSR-CNN.

\subsection{Examples}

Table \ref{tab:examples} shows example translations produced by both {\tt cdec} baselines, TSR-TXT, TSR-CNN, and TSR-HCA, together with source caption, image, and reference translation. The visual information induced by target side captions of pivot images allows a disambiguation of translation alternatives such as ``skirt'' versus ``rock (music)'' for the German ``Rock'', ``pole'' versus ``mast'' for the German ``Masten'', and is able to repair mistranslations such as ``foot'' instead of ``mouth'' for the German ``Maul''. 

\begin{table}
\begin{center}
\begin{footnotesize}
\begin{tabular}{|p{2cm} |p{4.8cm}|}

\hline
 Image: & \vspace{1pt} \includegraphics[scale=0.18]{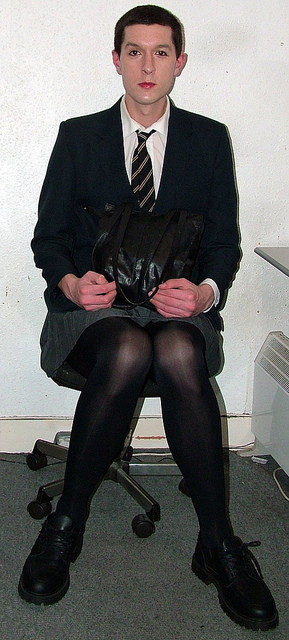}\vspace{3pt} \\
\hline
 Source:& Eine Person in einem Anzug und Krawatte und einem Rock.\\ 
\hline
{\tt cdec} out-dom: & a person in a suit and tie and a rock . \\
\hline
{\tt cdec} in-dom: & a person in a suit and tie and a rock .\\
\hline
 TSR-TXT:& a person in a suit and tie and a rock . \\ 
\hline
TSR-CNN: &  a person in a suit and tie and a skirt .\\
\hline
 TSR-HCA:&  a person in a suit and tie and a rock .  \\
\hline
 Reference:& a person wearing a suit and tie and a skirt\\          
\hline
\hline
Image: &\vspace{1pt} \includegraphics[scale=0.17]{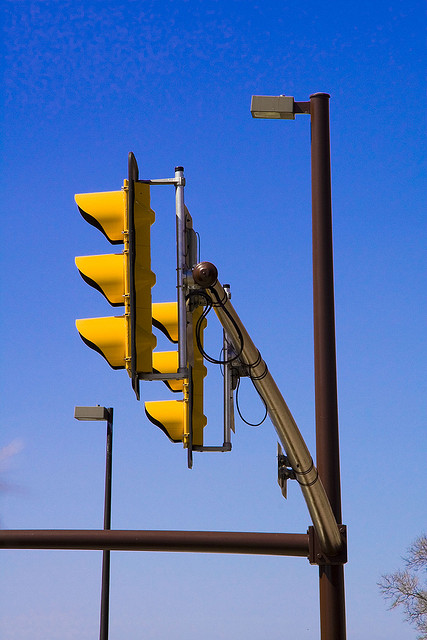}\vspace{3pt}\\
\hline
 Source:&  Ein Masten mit zwei Ampeln f\"ur Autofahrer. \\ 
\hline
{\tt cdec} out-dom: & a mast with two lights for drivers . \\
\hline
{\tt cdec} in-dom: & a mast with two lights for drivers .  \\
\hline
 TSR-TXT:& a mast with two lights for drivers .   \\ 
\hline
 TSR-CNN: & a pole with two lights for drivers .   \\
\hline
 TSR-HCA:& a pole with two lights for drivers .   \\
\hline
 Reference:& a pole has two street lights on it for drivers . \\          
\hline
\hline
Image: &\vspace{1pt} \includegraphics[scale=0.18]{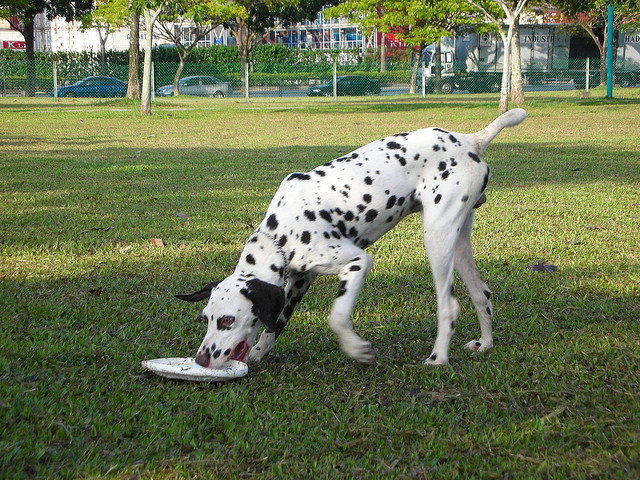}\vspace{3pt}\\
\hline
 Source:& Ein Hund auf einer Wiese mit einem Frisbee im Maul.\\ 
\hline
{\tt cdec} out-dom: & a dog on a lawn with a frisbee in the foot .\\
\hline
{\tt cdec} in-dom: & a dog with a frisbee in a grassy field .  \\
\hline
 TSR-TXT:& a dog with a frisbee in a grassy field .   \\ 
\hline
TSR-CNN: & a dog in a grassy field with a frisbee in its mouth .  \\
\hline
 TSR-HCA:&  a dog with a frisbee in a grassy field .    \\
\hline
 Reference:& a dog in a field with a frisbee in its mouth \\          
\hline

\end{tabular}
\end{footnotesize}
\end{center}
\caption{\label{tab:examples} Examples for improved caption translation by multimodal feedback.}
\end{table}

\section{Conclusions and Further Work}
We demonstrated that the incorporation of multimodal pivots into a target-side retrieval model improved SMT performance compared to a strong in-domain baseline in terms of BLEU, METEOR and TER on our parallel dataset derived from MS COCO. The gain in performance was comparable between a distance metric based on a deep convolutional network and one based on human object category annotations, demonstrating the effectiveness of the CNN-derived distance measure. Using our approach, SMT can, in certain cases, profit from multimodal context information. Crucially, this is possible without using large amounts of in-domain parallel text data, but instead using large amounts of monolingual image captions that are more readily available.

Learning semantically informative distance metrics using deep learning techniques is an area under active investigation \cite{Wu:2013:OMD:2502081.2502112,wang2014,Wang:2015:SSL:2764065.2764206}. Despite the fact that our simple distance metric performed comparably to human object annotations, using such high-level semantic distance metrics for caption translation by multimodal pivots is a promising avenue for further research.

The results were achieved on one language pair (German-English) and one corpus (MS COCO) only. As with all retrieval-based methods, generalized statements about the relative performance on corpora of various domains, sizes and qualities are difficult to substantiate. This problem is aggravated in the multimodal case, since the relevance of captions with respect to images varies greatly between different corpora \cite{Hodosh2013}. In future work, we plan to evaluate our approach in more naturalistic settings, such machine translation for captions in online multimedia repositories such as Wikimedia Commons\footnote{\url{https://commons.wikimedia.org/wiki/Main_Page}} and digitized art catalogues, as well as e-commerce localization.

A further avenue of future research is improving models such as that presented in \newcite{DBLP:journals/corr/ElliottFH15} by crucial components of neural MT such as ``attention mechanisms''. For example, the attention mechanism of \newcite{bahdanau2014neural} serves as a soft alignment that helps to guide the translation process by influencing the sequence in which source tokens are translated. A similar mechanism is used in \newcite{xu2015show} to decide which part of the image should influence which part of the generated caption. Combining these two types of attention mechanisms in a neural caption translation model is a natural next step in caption translation. While this is beyond the scope of this work, our models should provide an informative baseline against which to evaluate such methods.

\section*{Acknowledgments}
This research was supported in part by DFG grant RI-2221/2-1 ``Grounding Statistical Machine Translation in Perception and Action'', and by an Amazon Academic Research Award (AARA) ``Multimodal Pivots for Low Resource Machine Translation in E-Commerce Localization''.

\bibliography{literatur}
\bibliographystyle{acl2016}

\end{document}